\documentclass[letterpaper, 10 pt, conference, final]{ieeeconf}  % Use this line for final version

\IEEEoverridecommandlockouts                              % This command is only needed if 
\usepackage{graphicx}
\usepackage{subfigure}
\usepackage{caption}
\usepackage{color}
\usepackage{gensymb}
\usepackage{booktabs}
\usepackage[symbol]{footmisc}
\usepackage{ulem}
\usepackage{amsfonts}

\usepackage{multirow}
\usepackage{tabularx}
\usepackage[table]{xcolor}

\title{\LARGE \bf
T2FPV: Dataset and Method for Correcting First-Person View Errors in Pedestrian Trajectory Prediction
}

\author{Benjamin Stoler$^{1}$, Meghdeep Jana$^{2}$, Soonmin Hwang$^{3}$, and Jean Oh$^{3}$% <-this % stops a space
\thanks{$^{1}$Computer Science Dept.,
        Carnegie Mellon University, 5000 Forbes Ave, Pittsburgh, PA
        {\tt\small bstoler@cs.cmu.edu}}%
\thanks{$^{2}$ Mechanical Engineering Dept.,
        Carnegie Mellon University, 5000 Forbes Ave, Pittsburgh, PA
        {\tt\small mjana@andrew.cmu.edu}}%
\thanks{$^{3}$Robotics Institute,
        Carnegie Mellon University, 5000 Forbes Ave, Pittsburgh, PA%
        {\tt\small \{soonminh,jeanoh\}@cmu.edu}}%
}

\begin{document}

\maketitle

% See https://tex.stackexchange.com/questions/146803/place-a-two-column-picture-under-the-author-affiliation-and-above-abstract for reference
%\makeatletter
%\let\@oldmaketitle\@maketitle% Store \@maketitle
%\renewcommand{\@maketitle}{\@oldmaketitle% Update \@maketitle to insert...
  %\includegraphics[width=\linewidth]
    %{
    %figures/teaser_univ_wout_anno_v5}
    %\captionof{figure}{Top-down trajectories are replayed and recorded in a high-fidelity simulation; examples created from ETH/UCY~\cite{eth}}%
    %\vspace{-0.1cm}
    %\label{fig:teaser}%
    %}%
% perspectives, and the bottom is their corresponding segmentation view.}
%\makeatother\maketitle
%\setcounter{figure}{1}

%%%%%%%%%%%%%%%%%%%%%%%%%%%%%%%%%%%%%%%%%%%%%%%%%%%%%%%%%%%%%%%%%%%%%%%%%%%%%%%%
\begin{abstract} Predicting pedestrian motion is essential for developing socially-aware robots that interact in a crowded environment. While the natural visual perspective for a social interaction setting is an egocentric view, the majority of existing work in trajectory prediction therein has been investigated purely in the top-down trajectory space.
To support first-person view trajectory prediction research, we present T2FPV, a method for constructing high-fidelity first-person view (FPV) datasets given a real-world, top-down trajectory dataset; we showcase our approach on the ETH/UCY pedestrian dataset to generate the egocentric visual data of all interacting pedestrians, creating the T2FPV-ETH dataset. In this setting, FPV-specific errors arise due to imperfect detection and tracking, occlusions, and field-of-view (FOV) limitations of the camera. To address these errors, we propose CoFE, a module that further refines the imputation of missing data in an end-to-end manner with trajectory forecasting algorithms. Our method reduces the impact of such FPV errors on downstream prediction performance, decreasing displacement error by more than 10\% on average. To facilitate research engagement, we release our T2FPV-ETH dataset and software tools\footnote[4]{https://github.com/cmubig/T2FPV}.

%We report that the bird's-eye view assumption used in the original ETH/UCY dataset, i.e., an agent can observe everyone in the scene with perfect information, does not hold in the first-person views; only a fraction of agents are fully visible during each 20-timestep scene used commonly in existing work. 
%We evaluate existing trajectory prediction approaches under varying levels of realistic perception---displacement errors suffer a $356\%$ increase compared to the top-down, perfect information setting. 

\end{abstract}

\thispagestyle{empty}
\pagestyle{empty}

%%%%%%%%%%%%%%%%%%%%%%%%%%%%%%%%%%%%%%%%%%%%%%%%%%%%%%%%%%%%%%%%%%%%%%%%%%%%%%%%
\begin{figure*}[t]
    \centering    
    \vspace{2mm}
    \includegraphics[width=\linewidth]{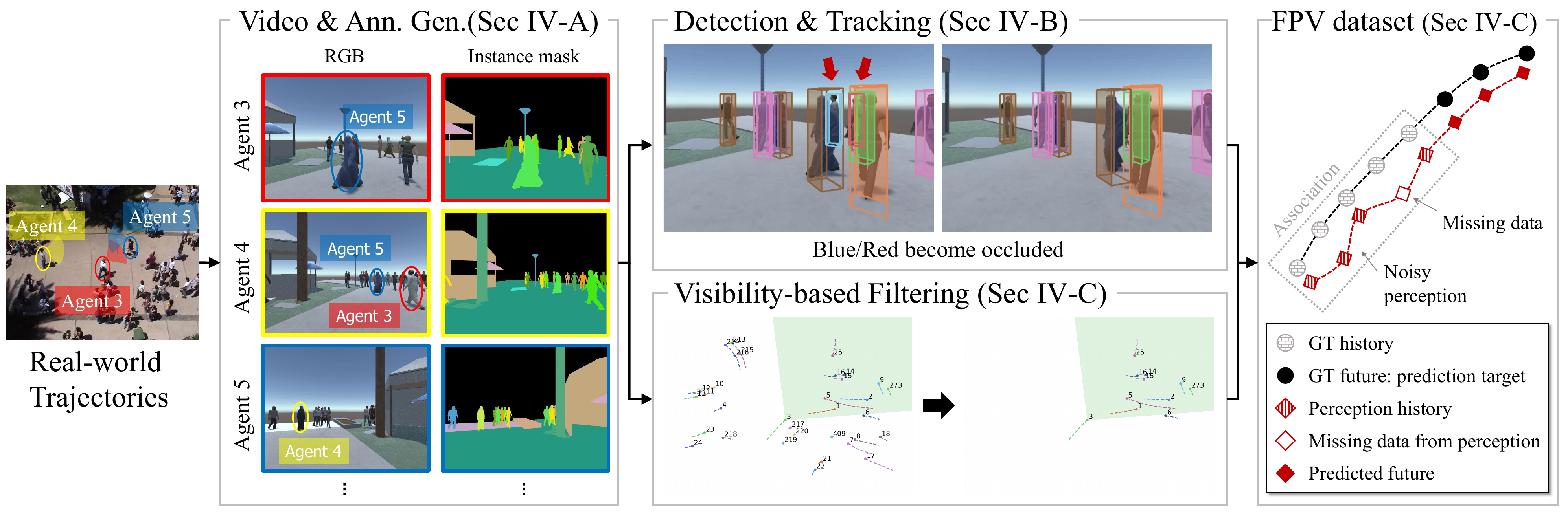}       
    
    \caption{\textbf{T2FPV Overview}: Wse generate filtered ground truth tracks and the corresponding D\&T tracks from a real-world pedestrian dataset. The downstream task is to predict the future path (black circles) for a set of perceived track histories (striped red diamonds).}
    % \vspace{-0.3cm}%
    \vspace{-2mm}
    \label{fig:overview}
\end{figure*}

\section{INTRODUCTION}

As more and more autonomous robots are anticipated to interact with people in shared environments, trajectory prediction in robotics has become increasingly popular in the research community, as well as among various industry and military stakeholders.
In particular, predicting pedestrian motion is essential for developing socially-aware robots that interact in a crowded environment~\cite{slstm,sgan,fpl,fvtraj,nsp}. Existing state-of-the-art (SOTA) trajectory prediction algorithms leverage datasets such as the ETH/UCY pedestrian dataset that provide full trajectory information of all pedestrians in a bird's-eye view (BEV) scene~\cite{eth}. 
However, bird's-eye view is an unrealistic view for agents navigating in the real-world; agents generally rely on egocentric, first-person view (FPV) sensing for these tasks. A realistic setting also includes limited field-of-view (FOV), occlusions, and changes in perspective and orientation of the ego-agent.

While collecting top-down data using an overhead camera is relatively convenient, creating a first-person view counterpart is far more challenging for several reasons. To begin with, all participants in the scene would need to wear a camera sensor to record their egocentric views, as well as a location-recording sensor to establish their ground truth locations. Furthermore, such a setting is subject to psychological issues such as the observer (or Hawthorne) effect~\cite{psych}, where people’s behaviors in these experiments may not be entirely representative of natural social interaction.

Therefore, to promote research on first-person view trajectory prediction, we propose T2FPV, a method for constructing an FPV version of data from a trajectory-only dataset by simulating the agents in high fidelity. 
%Each agent follows their recorded trajectory with a simulated camera attached to them. 
The FPV data is collected by having each agent follow their recorded trajectory with a simulated camera attached to them. 
%and we perform extensive annotation and post-processing to provide information beyond existing real-world first-person-view pedestrian datasets as well as previous synthetic first-person view datasets: 
We perform extensive annotation and post-processing to provide unique information beyond existing FPV datasets as follows:
%and we perform extensive annotation and post-processing to provide information beyond existing real-world first-person-view pedestrian datasets as well as previous synthetic first-person view datasets: 
1) we conduct SOTA detection-and-tracking, giving realistic partial perception of trajectories and enforcing data imputation as a core aspect of the task (in contrast with prior works which simply re-provide ground-truth BEV trajectories~\cite{fvtraj, deepsoc});  
2) our approach utilizes SEANavBench~\cite{sean}, a high-fidelity simulation environment, to provide realistic synthetic images; and 
%to provide a higher quality of synthetic images; and
3)  we additionally provide the corresponding ground truth of all observed and missed points of each trajectory, for its history and future (compared to only having information perceived from the camera as in ~\cite{fpl,krishnacam,qiu2022egocentric}). An overview of our approach is shown visually in Figure \ref{fig:overview}.

%\todo{}{Consider shortening the previous list?}

To showcase our approach, we construct the T2FPV-ETH dataset based on the ETH/UCY trajectory dataset~\cite{eth}. In this realistic FPV setting, we observe that a new class of errors is present compared to in BEV. These ``FPV errors'' arise from occlusion and field-of-view (FOV) limitations of robot sensing, combined with imperfect detection and tracking, resulting in missing observations. When performing trajectory prediction with various SOTA approaches, these errors caused our observed metrics to be significantly worse than what was reported in the BEV setting in the literature\footnote{For instance, Average Displacement Error (ADE) / Final Displacement Error (FDE) performance increased from 0.44m / 0.89m in BEV to 1.51m / 2.08m in FPV for VRNN \cite{acvrnn}}. 

%(e.g., SGNet performance was 1.00 / 1.28 with linear interp, compared with their reported numbers of 0.18 / 0.35 \cite{sgnet}).

Prior work in pedestrian prediction has largely ignored FPV errors, either throwing out incomplete tracks or relying on simple interpolation over the missing points~\cite{retrack}~\cite{weng2022whose}. Recent work has made significant advancement in data imputation; however, the vast majority of these works focus on 
%a synthetic setting where data is made to be artificially missing
artificially missing data~\cite{zhao2022trajgat, naomi19, 2020both}. Additionally, these works only focus on imputation as an independent task without considering how it affects prediction performance. Hence, to reduce the FPV errors for improved prediction, we propose \textbf{Co}rrection of \textbf{F}PV \textbf{E}rrors (CoFE) that can refine initial imputations via end-to-end training with a trajectory prediction approach. 
%We find that we are able to decrease prediction displacement errors by approximately 12\% each when using this approach on SOTA imputation and forecasting approaches.
We find that our approach decreases prediction displacement errors by more than 10\% on average when compared to all tested imputation and forecasting combinations.

%\todo{}{This is just the average of ADE and FDE improvements over all 3 algos x all 3 imputations, i.e. 11.5\% and 12\%, is that what we should include here? An alternative view is to look at linear-interpolation as the de-facto baseline, since that is what used typically in our field..} 

%\todo{}{ALSO, how novel should we claim this is? e.g. ``To the best of our knowledge, this is the first approach to perform combined imputation and forecasting on human motion trajectories?'' like can we claim that at all? }

Our main contributions are: 
1) we propose a method for creating an egocentric view for each agent given a set of trajectories; 
2) we generate the T2FPV-ETH dataset, a new first-person view dataset that corresponds to the ETH/UCY dataset; 
3) we propose and evaluate CoFE, an end-to-end learned input correction module, which reduces the impact of FPV errors beyond SOTA imputation approaches; and
4) we release our dataset and software tools to promote research in first-person view trajectory prediction.

\section{RELATED WORK}
\label{sec:rel_work}

\textbf{Real-World First-Person Datasets.} Various large-scale datasets provide video footage from an ego agent's perspective. \cite{ego4D} is a large-scale first-person view video dataset, with over 3500 hours of footage collected from various sources around the world.  Egocentric Basketball Motion Planning \cite{bball} provides a wearable camera perspective from multiple people in the scene. However, neither of these datasets are focused on social navigation. They feature many instances of the ego agent walking by themselves or performing an unrelated task (such as carpentry, basketball, etc.) that have inherently different social contexts than navigating in public. 

\cite{krishnacam} is a dataset of an egocentric pedestrian video stream, providing pose, acceleration, and orientation information. Similarly, \cite{fpl} uses human camera wearers and pose estimation to create a dataset of 2D targets to predict. However, these approaches only provide a single perspective of an ego agent in each scenario, limiting the diversity of ego behaviors. Also, both lack the ground truth pose information of other agents in the scene, especially due to occlusion and FOV limits. 

Self-driving datasets, such as \cite{waymo, caesar2020nuscenes}, suffer the same problem of not having full ground truth information for training and evaluation. Furthermore, a car's ego-motion is incomparable to a pedestrian's ego-motion in terms of physical characteristics. Additionally, the social interactions and roles between the ego and detected agents are vastly different between the two fields.

\textbf{Synthetic Pedestrian Datasets and Simulation.} Several recent works have generated synthetic data in simulations based on a corresponding real-world dataset. FvTraj~\cite{fvtraj} uses Unity to render FPV images from ground truth trajectory data \cite{unity}, but these rendered images consist only of a flat ground plane with no corresponding environment modeled. DeepSocNav~\cite{deepsoc} generates ego view depth images from ETH/UCY, with a low-fidelity environment model. However, they do not include images from RGB cameras, which are far more common than depth sensors. Furthermore, DeepSocNav~\cite{deepsoc} and FvTraj~\cite{fvtraj} do not release any generated images or their in-house simulators, limiting reproducibility and engagement within the research community. Most importantly, both works only use the generated images for augmenting ground truth trajectories when performing prediction; no perception or detection and tracking is used, keeping the task less realistic.

\cite{garden} and \cite{sean} are relatively high-fidelity simulation environments with scene constructions of ETH/UCY, built in Unreal Engine~\cite{unrealengine} and Unity~\cite{unity} respectively, but both lack first-person views. Furthermore, these approaches also have the same aforementioned limitations as \cite{deepsoc, fvtraj} regarding perception and task settings.

\textbf{Pedestrian Trajectory Prediction.} Recent work on trajectory prediction and forecasting has mostly focused on top-down trajectory datasets such as ETH/UCY~\cite{eth}, SDD~\cite{sdd}, and inD~\cite{ind}. \cite{slstm} uses LSTMs to jointly predict trajectories of all agents, incorporating pooled hidden-state information from neighbors as a social cue. Some approaches, such as AC-VRNN~\cite{acvrnn},  use generative models within a VRNN~\cite{vrnn}, incorporating social interactions via attentive hidden state refinement. 
Several works also leverage top-down images explicitly, whether in an RGB form or with added semantic segmentation~\cite{nsp, ynet, v2net}. SGNet~\cite{sgnet} generates coarse step-wise goals to assist trajectory prediction sequentially. \cite{trajectron} incorporates agent dynamics and environment information and forecasts using a graph-structured recurrent model. 

Fewer works have focused on the FPV setting for pedestrians. \cite{fpl} utilizes FPV to model and predict the trajectory of a single agent directly in pixel-space. \cite{park2016} creates a spatial visual distribution of objects from FPV, and applies perception and ego-agent trajectory planning in a 2.5D coordinate system. \cite{qiu2022egocentric} uses a transformer-based architecture, with a graph scene encoding to forecast the camera wearer's trajectory with nearby agents as a cue. Still, none of these works deal with FPV errors when providing the trajectories of detections.

\textbf{Trajectory Robustness.}
The field of sequence imputation has had much success with deep learning recently. NAOMI~\cite{naomi19} uses a non-autoregressive approach at multiple step sizes to impute missing data in the context of basketball players and billiard ball trajectories. \cite{2020both} trains imputation and prediction together but still only evaluates them separately rather than as an end-to-end pipeline. \cite{ingrain}  evaluates the end-to-end task but only focuses on low-resolution, long-term GPS data, dissimilar to the fine-grained social navigation task. Also, these approaches only deal with artificial missing-completely-at-random (MCAR) data rather than dealing with pathologically missing data due to FPV errors.

\cite{zhao2022trajgat} focuses on real vehicle and human motion trajectories, by transforming existing forecasting challenges into imputation ones. However, similar to the above approaches, they still only apply MCAR masks to ground truth. Furthermore, they only deal with imputation between ground truth points, rather than points which themselves may be erroneous from the perception model.

There have been several recent works in improving the robustness of trajectory forecasting to perception errors. \cite{retrack} combines refinement via exponential smoothing with trajectory prediction to iteratively re-match observed trajectories with ground truth. \cite{weng2022whose} reframes the perception pipeline to remove tracking altogether, instead operating directly on detections and affinity matrices. However, both approaches still rely only on simple linear interpolation and extrapolation for missing data. While these are interesting approaches that we believe to be complementary to ours, we leave them as future work as they primarily focus on tracking and data association errors.

\section{PROBLEM FORMULATION}
\label{sec:prob-formulation}

A trajectory prediction problem using complete information is defined as follows: for $N$ pedestrians in a scene, we denote the position of each agent $i$ in the $xy$ ground-plane at time-step $t$ as $\mathbf{X_i^t} = (x_i^t, y_i^t)$. Given the observed track histories,  $\mathbf{X_i^{hist}} = \{\mathbf{X_i^t} | t=1,2,...,T_{obs}\}$, the task is to predict the future paths $\mathbf{X_i^{fut}} = \{\mathbf{X_i^t} | t=T_{obs+1},T_{obs+2},...,T_{pred}\}$ for all agents $i$ in a given scene, including the ego agent.

In this paper, we introduce a trajectory prediction task where each agent is to predict the trajectories of all agents in their view only using their egocentric information. Hence, observed non-ego track histories may be erroneous compared to the ground truth (GT). 
Thus, given that $\mathbf{X_{i}^t}$ denotes the ground truth position of a given agent $i$ at time $t$, we will define ${\mathbf{\tilde{X}_{i}^t}}$ to be the estimated position of agent $i$ at time $t$. Similarly, ${\mathbf{\tilde{X}_{i}^{hist}}}$ will represent an estimated history portion of the agent, where missing points have been imputed by some method. 
If the scene has $N$ agents, note that there is a single ego agent $e$, and $N-1$ detected agents. Thus, the FPV trajectory prediction problem involves predicting $\mathbf{X_i^{fut}}$ for each agent $i$ in a scene, given $\mathbf{\tilde{X}_{d_i}^{hist}}$ for each detected agent $d_i$ and $\mathbf{X_e^{hist}}$ for the ego agent, $e$.

%Formally, let $\phi_{i,t}$ denote agent $i$'s FPV image at time step $t$ and $\Psi(i)$ denote the set of agents that are within agent $i$'s field of view. Then, agent $j \in \Psi(i)$ if agent $i$'s views $\{\phi_{i, t}| t=t',t'+1,...,t'+k\}$ contain at least some $P$ pixels associated with agent $j$, for some $k$ time steps. For each agent $i$ in $1,...,N$, the FPV trajectory prediction task is to predict the future trajectories of agent $i$ and all agents within agent $i$'s FOV, given FPV observations, $\{\phi_{i, t}| t=1,...T_{obs}\}$, as well as their ego track history.

\section{TRAJECTORIES TO FIRST-PERSON VIEW}
\label{dataset_creation}

We describe our T2FPV method, demonstrating how we construct first-person view data from an example trajectory dataset, namely the ETH/UCY dataset. This dataset consists of five ``folds'' of recorded data, in different locations and times: ETH, Hotel, Univ, Zara1, and Zara2.

\subsection{Video and Annotation Generation}

Our approach for creating FPV datasets from real-world trajectory datasets begins with generating videos and ground-truth annotations. We use the SEANavBench~\cite{sean} simulation environment as a starting point for our simulation. SEANavBench consists of high-fidelity pre-modeled scenes for each location within ETH/UCY. We leave these scenes as unchanged as possible, for consistency with prior works using SEANavBench.

As in~\cite{fvtraj}, we enforce a number of assumptions when rendering these tracks. For instance, we orient each pedestrian's gaze with the direction they are traveling in, with spherical linear interpolation for smoother angle changes. Additionally, we mount a camera on each pedestrian at a fixed height of $1.6m$ from their base and assign the following physical characteristics to the camera: $18mm$ focal length, $36\times24mm$ sensor, and zero lens shift for the principal point. When rendered at our $640\times480$ resolution, this results in a vertical FOV of approximately $67\degree$. 

Using the above assumptions, we then render the first-person videos for every person following their track from the original dataset, as well as output an annotation for each agent at every frame. The videos consist of the RGB render, as well as an instance segmentation render, as shown in Figure \ref{fig:overview}, where each object in the scene has been given a unique color. The annotations consist of the agent's ID, pose information, and a list of what other agents can be seen in the camera's view, i.e., the poses of all visible agents in both the camera and world reference frame. This detection list is generated by utilizing the aforementioned segmentation mask to determine agent visibility.  

\subsection{Perception: Detection and Tracking}
\label{det_track_desc}
%\soonmin{"Perception tracks from Detection and Tracking" for consistency to Sec 4-C?}
% % \begin{figure*}[t]
% \begin{figure}[t]
%     \centering
%     \vspace{0.4cm}
%     \subfigure[Groundtruth]{
%         % \includegraphics[scale=0.2]{figures/DnT_examples/crowds_zara01-a0056-f003740___viz_gt_boxes3d_cam_track___2crhlpec___model_final.png}    
%         \includegraphics[width=0.45\linewidth]{figures/DnT_examples/crowds_zara01-a0056-f003740___viz_gt_boxes3d_cam_track___2crhlpec___model_final.png}    
%     }
%     \subfigure[Prediction]{
%         % \includegraphics[scale=0.2]{figures/DnT_examples/crowds_zara01-a0056-f003740___viz_pred_boxes3d_cam_track___2crhlpec___model_final.png}
%         \includegraphics[width=0.45\linewidth]{figures/DnT_examples/crowds_zara01-a0056-f003740___viz_pred_boxes3d_cam_track___2crhlpec___model_final.png}
%     }
%     % \subfigure[Top-down view (green: groundtruth, blue: prediction)]{
%     %     \includegraphics[scale=0.2]{figures/DnT_examples/crowds_zara01-a0056-f003740___viz_boxes3d_bev_track___2crhlpec___model_final.png}
%     % }
%     \caption{Example of our detection and tracking results on Zara1; pedestrians are assigned colors based on their tracked ID. 
%     % The left image represents the ground truth 3D bounding box and instancing, while the middle is the predicted results. The right image shows the overlay of these bounding boxes from a top-down perspective, with concentric circles at 5m intervals from the camera.
%     }\vspace{-0.3cm}%
%     \label{fig:det_track}
% % \end{figure*}
% \end{figure}

To perform trajectory prediction in a realistic setting, we employed an off-the-shelf object detector and tracker to produce the observations required. We used a 3D object detector~\cite{park2021dd3d} which is SOTA among recent image-only methods which do not require depth information~\cite{ma20223d}, and a simple but effective probabilistic tracker~\cite{kuang2020probabilistic}. We made the following changes to both approaches to produce reasonable detection and tracking results.

In DD3D~\cite{park2021dd3d}, we set the parameters of feature map assignment to use thresholds that fit our ground truths appropriately. We also only used instances that are ``visible'' (as defined in Section \ref{sec:fpv_gt}), which helps to filter out heavily occluded instances. For the tracker~\cite{kuang2020probabilistic}, we changed the matching metric to use BEV IoU (Intersection-over-Union in top-down view) from Mahalanobis distance~\cite{mahalanobis} to associate detections to tracks. We also applied the Kalman filter only to each instance's 3D location and orientation and used state and observation noise covariances calculated from our ground truth data.

Following the common evaluation procedure as in the ETH/UCY trajectory prediction task, we trained one model for each of the five folds, using the other four folds as the training and validation sets respectively. We then produced tracking results on all ego videos from each fold's test-set.

\begin{table}[t]
\vspace{2mm}
\caption{Detection and tracking performance.}
\label{table:dnt_metrics}
\definecolor{lightgray}{gray}{0.95}
\newcolumntype{Y}{>{\centering\arraybackslash}X}
\begin{tabularx}{\linewidth}{c|YY|YY}
\toprule
 \multirow{2}{*}{Fold} & \multicolumn{2}{c|}{Detection} & \multicolumn{2}{c}{Tracking} \\
% \cline{2-5}
 & $AP_{2D}$ & $AP_{BEV}$ & AMOTA & AMOTP \\
\midrule
\rowcolor{lightgray} 
ETH     & 96.50 & 44.10 & 0.384 & 1.262 \\
Hotel   & 94.24 & 42.56 & 0.361 & 1.325 \\
\rowcolor{lightgray} 
Univ    & 90.65 & 67.56 & 0.318 & 1.465 \\
Zara1   & 97.29 & 90.22 & 0.709 & 0.610 \\
\rowcolor{lightgray} 
Zara2   & 94.67 & 73.78 & 0.517 & 1.000 \\
\bottomrule
\end{tabularx}
\vspace{-2mm}
\end{table}

\subsection{FPV Dataset Creation}
\label{sec:fpv_gt}

In transitioning from \textit{bird's-eye-view} (BEV) to \textit{first-person view} (FPV), given a scene with $N$ agents, we now construct $N$ variations of the original scene, i.e., from each agent's perspective. We begin with the same pre-processing popularized in Social GAN~\cite{sgan}, only considering scenes with at least two concurrent agents in a sliding window consisting of $T_{obs} = 8$ and $T_{pred} = 12$ time steps.
Then, to account for FPV errors, we redesign the scene as follows, for each agent's perspective.

First, we consider the set of observed tracks from the detection and tracking module. We filter out tracks that are seen by the ego agent for fewer than $k$ of the first $T_{obs}$ time steps. Next, we perform an initial imputation on missing values using linear interpolation (as in \cite{weng2022whose}). This creates a D\&T set of tracks, consisting of $\mathbf{\tilde{X}_{{d_i}'}^{hist}}$ for each detection.

We then consider the set of ground truth tracks from BEV. We filter out tracks that are impossible to have been seen by the ego agent for fewer than $k$ of the first $T_{obs}$ time steps (i.e. by having fewer than $P$ pixels visible from instance segmentation). Furthermore, we filter out tracks for which the ground truth is missing pieces of data for any of $T_{obs}$ or $T_{pred}$. In our creation of T2FPV-ETH, we used $k=3$ and $P=100$. This step then creates a GT set of tracks, consisting of $\mathbf{X_{d_i}^{hist}}$ as well as $\mathbf{X_{d_i}^{fut}}$ for each agent which is feasibly visible to the ego agent.

For each scene, the GT and detected sets of tracks are associated together by performing Hungarian matching, as in \cite{retrack, weng2022mtp, weng2022whose}, based on the mean squared error (MSE) between each $\mathbf{X_{d_i}^{hist}}$ and $\mathbf{\tilde{X}_{{d_k}}^{hist}}$. Finally, each scene has the corresponding ego agent $\mathbf{X_e^{hist}}$ and $\mathbf{X_e^{fut}}$ appended.

\subsection{Dataset Statistics}
\label{sec:stats}

\begin{table}[t]
\centering
\vspace{2mm}
\caption{T2FPV-ETH statistics.} 
\definecolor{lightgray}{gray}{0.95}
\begin{tabular}{c|cccc}
\toprule
Fold & Num Ego & Num Dets & Det MSE & FPV Err. Rate \\
\midrule
\rowcolor{lightgray} 
ETH & 181 & 60 & 2.05 & 0.44 \\
Hotel & 1053 & 449 & 2.03 & 0.51\\
\rowcolor{lightgray} 
Univ & 24,334 & 120,072 & 1.13 & 0.45 \\
Zara1 & 5,939 & 3,686 & 0.64 & 0.28 \\
\rowcolor{lightgray} 
Zara2 & 17,608 & 11,775 & 1.05 & 0.32 \\
\bottomrule
\end{tabular}
\label{tab:simple_stats}
\vspace{-2mm}
\end{table}

We measure the detection and tracking performances of the SOTA methods we employed in Table \ref{table:dnt_metrics}. 
For detection performance, we measure the standard average precision ($AP_{2D}$) in 2D image space and observe that it performs well. Also, we measure the localization quality of detected objects in 3D space by calculating IoU-based average precision in the top-down view ($AP_{BEV}$). Both metrics use the same IoU threshold of 0.5. The $AP_{BEV}$ performance is worse than $AP_{2D}$, which shows the challenge of image-based 3D detection.
For tracking, we adopt two popular metrics from \cite{weng20203d}, Average Multi-Object Tracking Accuracy (AMOTA) and Precision (AMOTP). AMOTA combines false positives, missed targets, and identity switches, and AMOTP measures the misalignment between prediction and ground truth. Although ``Univ" shows the worst performance because of the pedestrian density (Table \ref{tab:simple_stats}), the detector and tracker perform reasonably well, as shown qualitatively in Figure \ref{fig:overview}.

Table \ref{tab:simple_stats} provides a high-level overview of the number of scenes and detections, as created in Section \ref{sec:fpv_gt}. We note that this table demonstrates a data augmentation effect, as there is now a one-to-one correspondence between each ego agent and a scene; a single ground-truth track is often observed by multiple other agents at once, although with different possible FPV errors and 3D detection locations. These statistics indicate the diversity between the different folds as testing sets, as they have significantly varying scene densities (i.e. detections per ego agents), as well as rate of FPV errors (i.e. number of points needing to be imputed downstream) and difficulty of imputation.
\begin{figure*}[t]
    \centering
    \vspace{2mm}
    \includegraphics[width=0.98\linewidth]{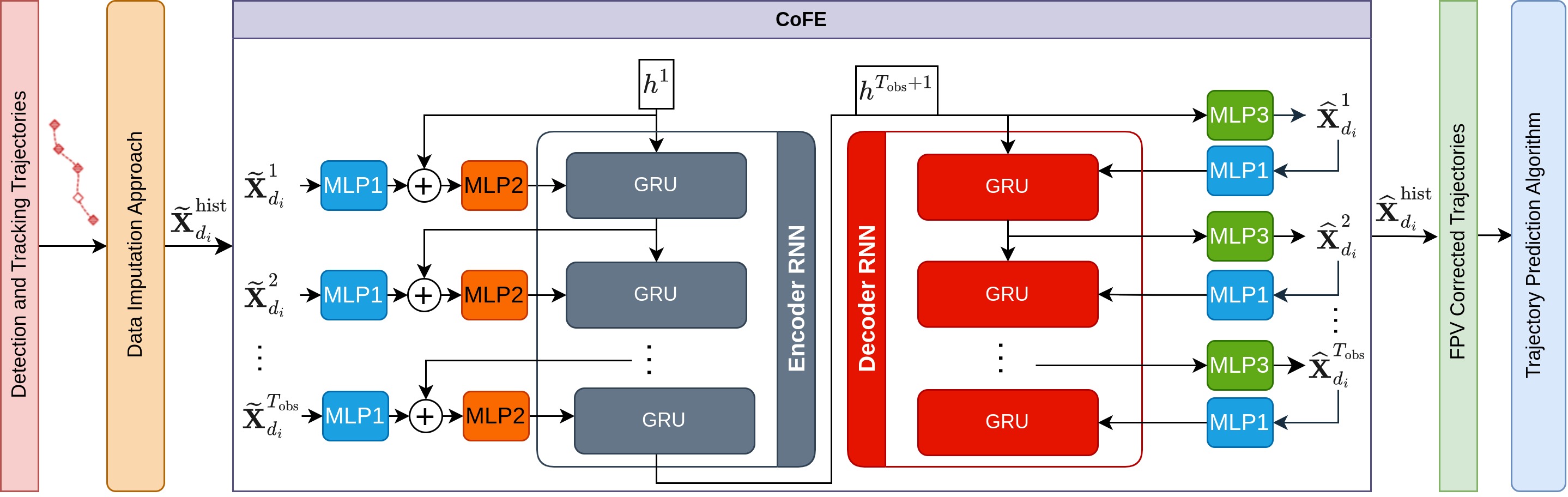}
    \caption{\textbf{CoFE Approach}: CoFE has an encoder-decoder architecture that refines the imputed trajectories to better account for FPV errors. The corrected trajectories are then passed into a trajectory prediction algorithm. }\vspace{-0.3cm}%
    \label{fig:arch}
\end{figure*}

\section{PROPOSED METHOD: CoFE}

%\soonmin{Explain why we need this, especially when we move to FPV setting.}
%\todo{}{Cite the dataset statistics for justification of how prominent these errors are and such.}

%\todo{}{Outline: 1. Quick paragraph on motivation/research gap, i.e. masking assumptions + trusting surrounding points
%2. Design to address it: incorporate existing SOTA imputation + RNN whole trajectory enc/dec + E2E training. Design choice still of just refining imputation rather than adjusting surrounding points, empirically it worked better
%3. Discussion about incorporation with trajectory prediction approaches, how they require full data in the first place.}

\subsection{Motivation}
\label{sec:Motivation}

As noted in Section \ref{sec:rel_work}, existing imputation approaches have two primary deficits when being applied to the field of human trajectory prediction. First, approaches largely use a missing-completely-at-random (MCAR) treatment of the points to be imputed. This assumption does not hold in a setting with FPV errors, as data is missing in a manner pathological to the detection and tracking approach being used as well as compulsory to occlusion and FOV limitations from the ego camera. Second, approaches have full trust in the accuracy of the points around the missing data. This assumption clearly also does not hold in the FPV setting, as the positions of observations are estimated from our approach described in Section \ref{det_track_desc}.

Therefore, we propose to incorporate a correction module, \textbf{Co}rrection of \textbf{F}PV \textbf{E}rrors (CoFE), between existing imputation approaches and downstream trajectory prediction approach, as shown in Figure \ref{fig:arch}. 
%This should be done in a way that is consistent with our downstream trajectory prediction task. Thus, we learn both the refinement of the imputed inputs as well as the trajectory prediction algorithm itself in an end-to-end (E2E) neural network. 
To make the correction consistent with the trajectory prediction task, we thus train a neural network for both the imputation refinement and the trajectory prediction algorithm itself in an end-to-end (E2E) manner. 

\subsection{CoFE Architecture}
\label{sec:architecture}

As shown in Figure \ref{fig:arch}, our architecture is similar to many RNN-based trajectory forecasting approaches, such as VRNN~\cite{vrnn}. Our main insight is to use an encoder RNN and decoder RNN back to back. The accumulated hidden state after processing the input then feeds into the decoder to output a ``corrected'' version of the input. We utilize two different Gated Recurrent Unit (GRU) modules for this purpose.

For a detected agent $d$, each missing point of $\mathbf{\tilde{X}_{d_i}^{hist}}$ is imputed via the given approach, e.g., NAOMI~\cite{naomi19}. Then, each point is transformed into its relative motion from the previous point, in order to be agnostic to absolute coordinates in the given scene. Next, these relative motions are feature extracted with a Multilayer Perceptron (MLP), labeled \texttt{MLP1} in our diagram. These features are concatenated with the hidden state $h^{t}$ at each time step in $T_{obs}$, then fed into \texttt{MLP2} and the encoding GRU cell to obtain the next hidden state, $h^{t+1}$. After processing the entire input, we then switch to decoding with the last hidden state, $h^{T_{obs}+1}$. We output $\mathbf{\hat{X}_{d_i}^{1}}$ as a prediction for $\mathbf{X_{d_i}^1}$, via \texttt{MLP3}. We then apply the same feature extractor (i.e., same weights) \texttt{MLP1} to this prediction to then feed back into the decoding GRU cell, and repeat this process for all $T_{obs}$ points. Finally, we convert the predicted points back into absolute coordinates. The exact details of this architecture, including the number of layers and hidden units in each MLP, can be seen in our open-sourced implementation. 

\subsection{End-to-End (E2E) Training}
\label{sec:training}

We introduce a simple MSE Loss objective to train CoFE itself, between the ground truth $\mathbf{X_{d_i}^{hist}}$ and corrected estimations $\mathbf{\hat{X}_{d_i}^{hist}}$. Given the original estimated points with imputation $\mathbf{\tilde{X}_{d_i}^{hist}}$ and also the refinements $\mathbf{\hat{X}_{d_i}^{hist}}$, we then update the points in $\mathbf{\hat{X}_{d_i}^{t}}$ with $\mathbf{\tilde{X}_{d_i}^{t}}$ for timesteps $t$ where imputation is not required. This final $\mathbf{\hat{X}_{d_i}^{t}}$ is then used to train the downstream prediction method (e.g., SGNet~\cite{sgnet}) in an end-to-end (E2E) manner, where the Loss function being optimized is the sum of the CoFE Loss objective and the prediction method's original objective.

\begin{figure*}[t]
    \centering
    \vspace{2mm}
    \subfigure[Scenario 1 (From Zara2)]{
        \includegraphics[width=0.32\textwidth]{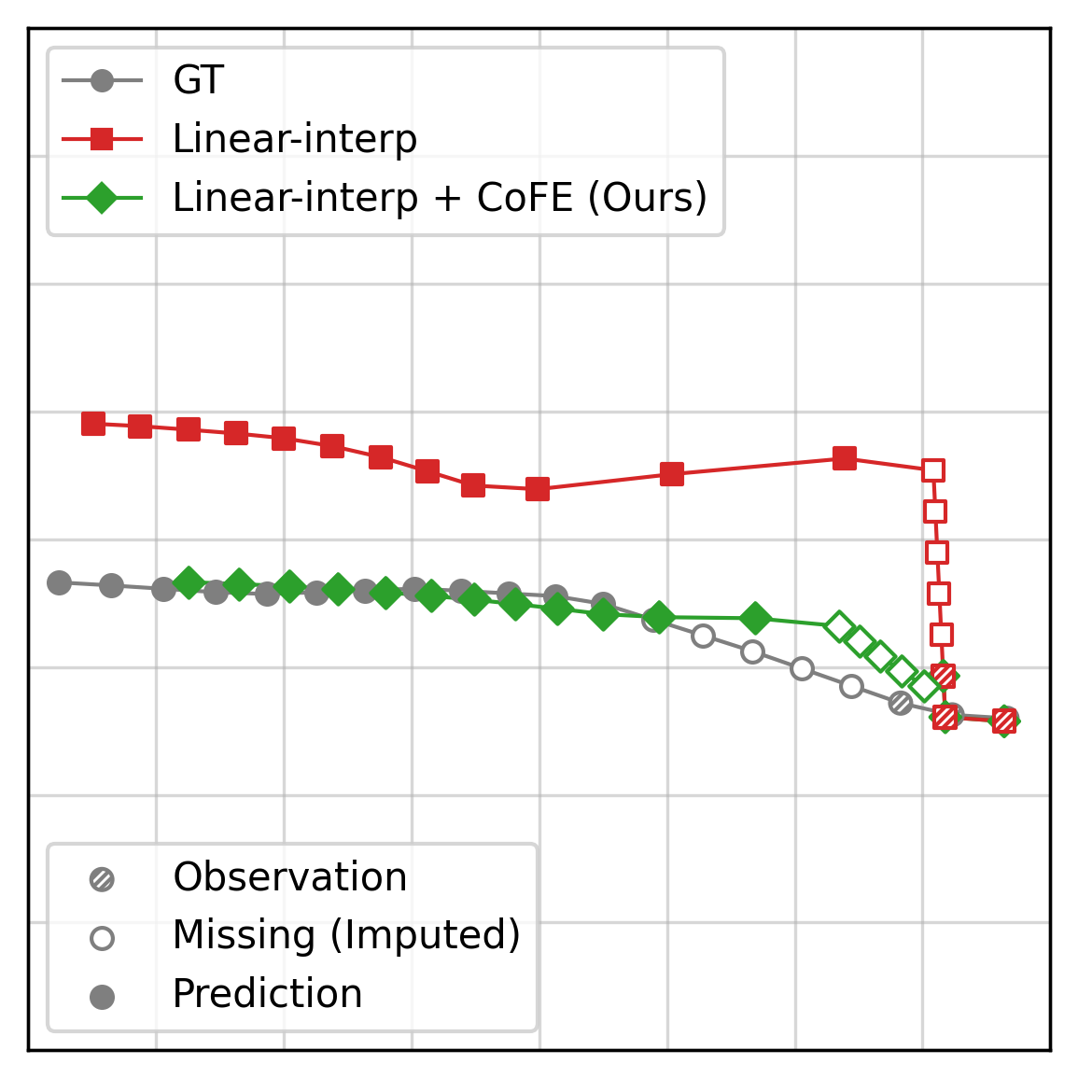}   
        \includegraphics[width=0.32\textwidth]{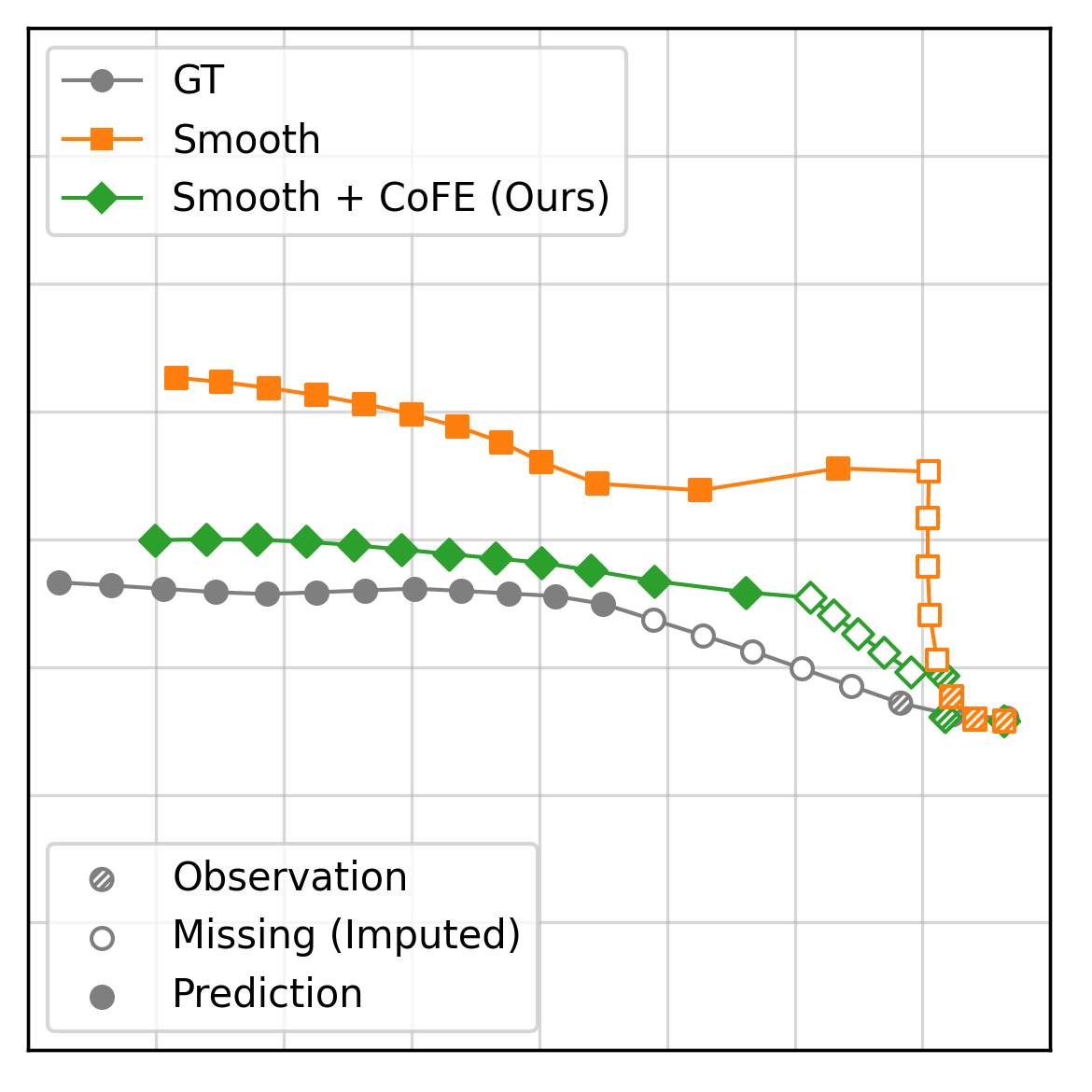}   
        \includegraphics[width=0.32\textwidth]{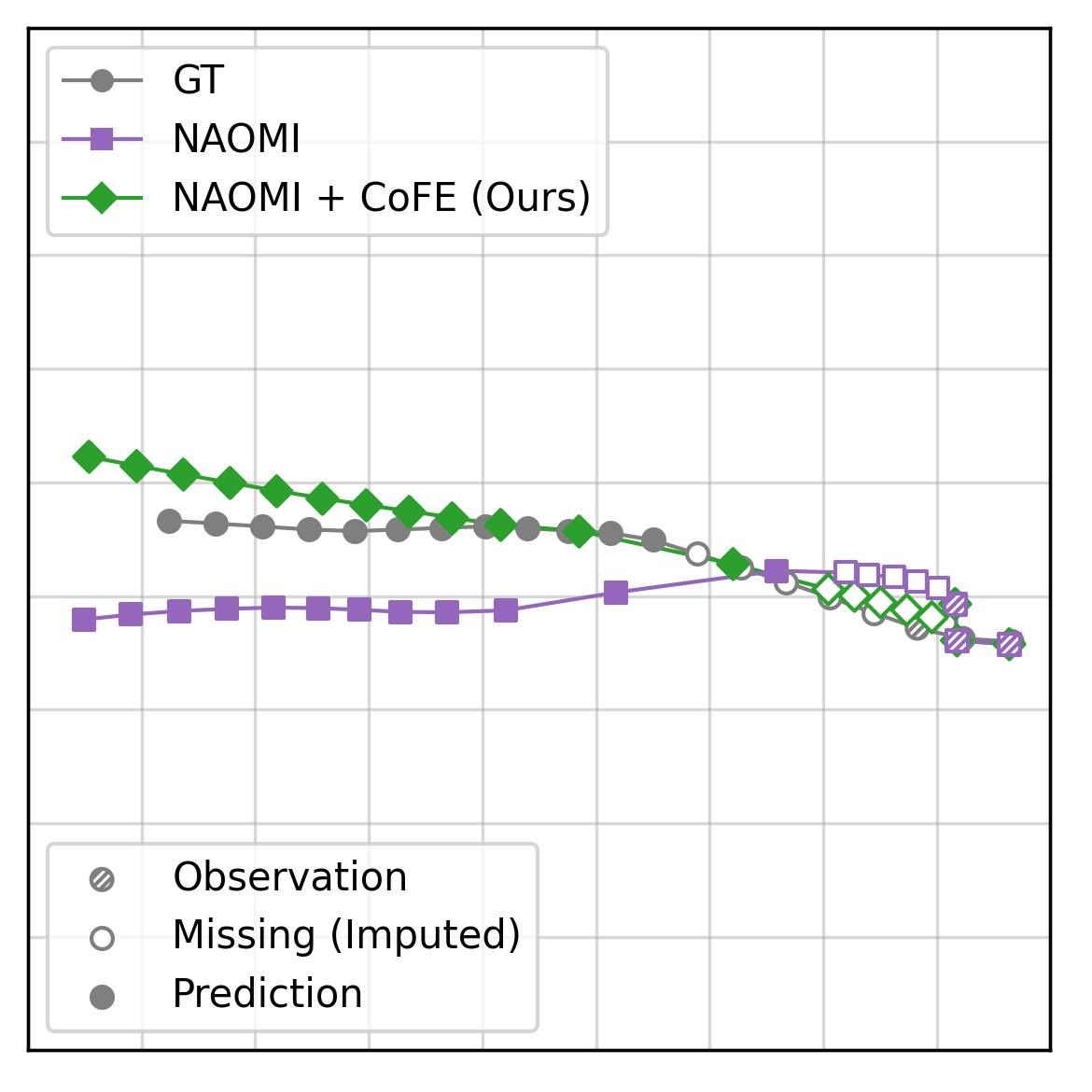}           
    }

    \subfigure[Scenario 2 (From Univ)]{
        \includegraphics[width=0.32\textwidth]{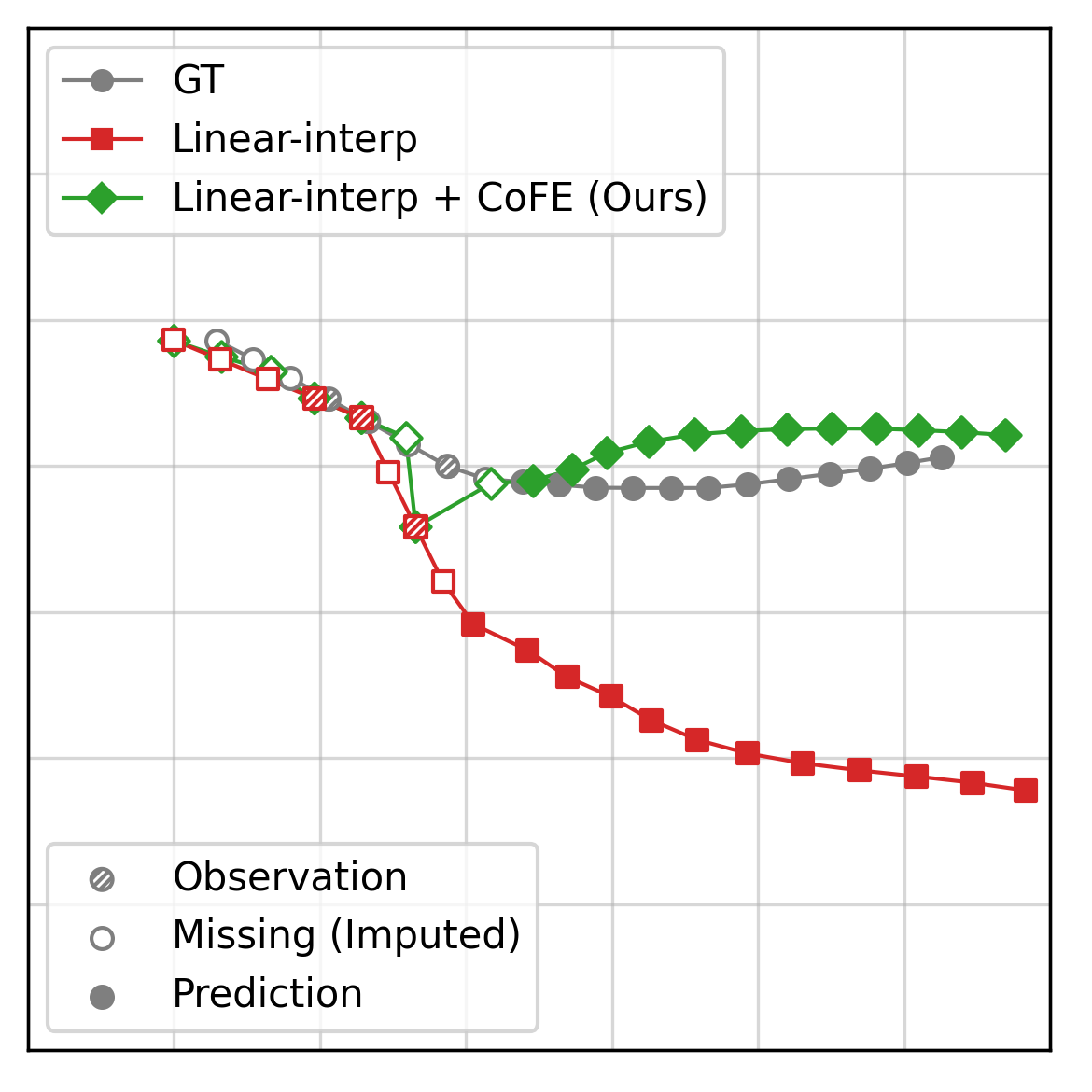}   
        \includegraphics[width=0.32\textwidth]{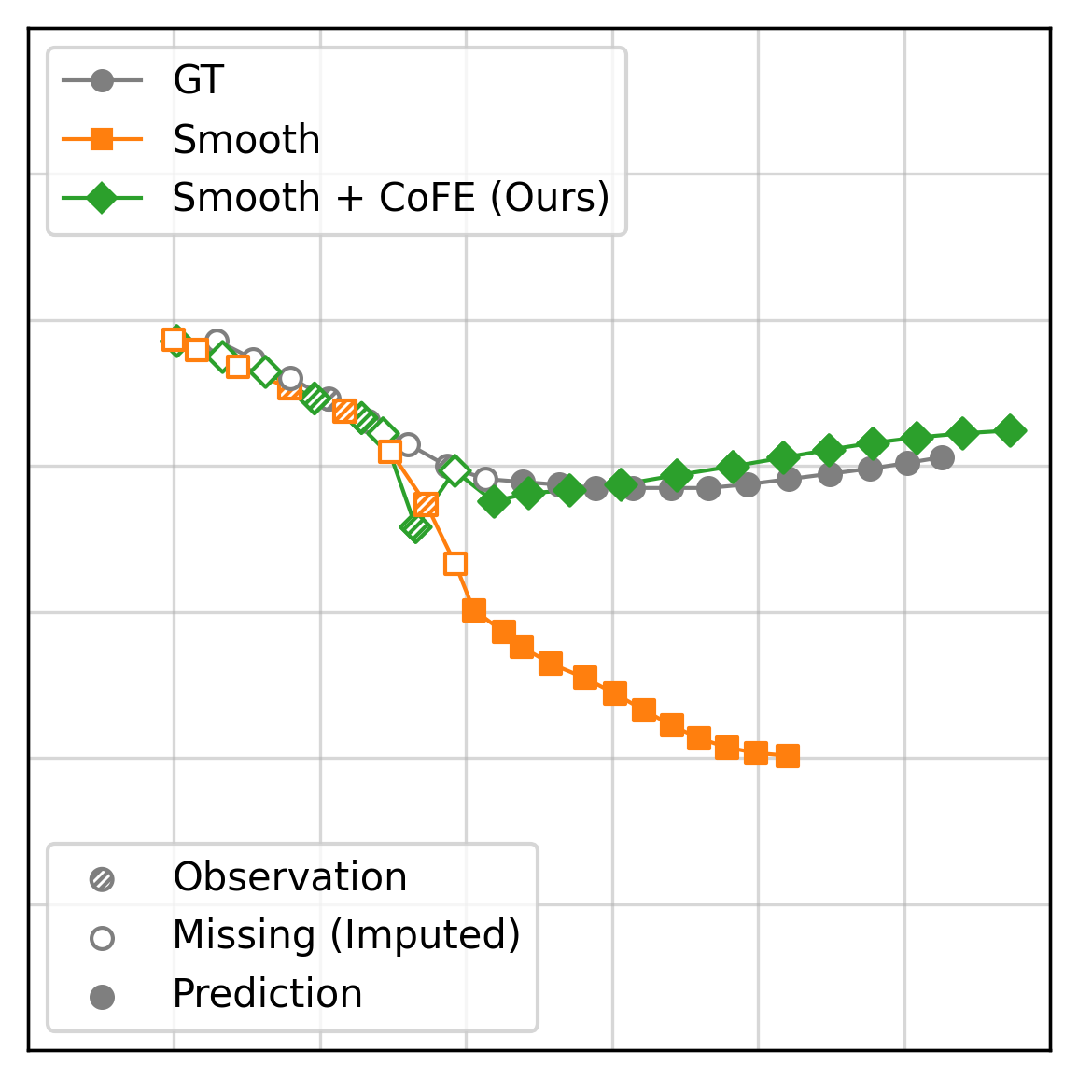}   
        \includegraphics[width=0.32\textwidth]{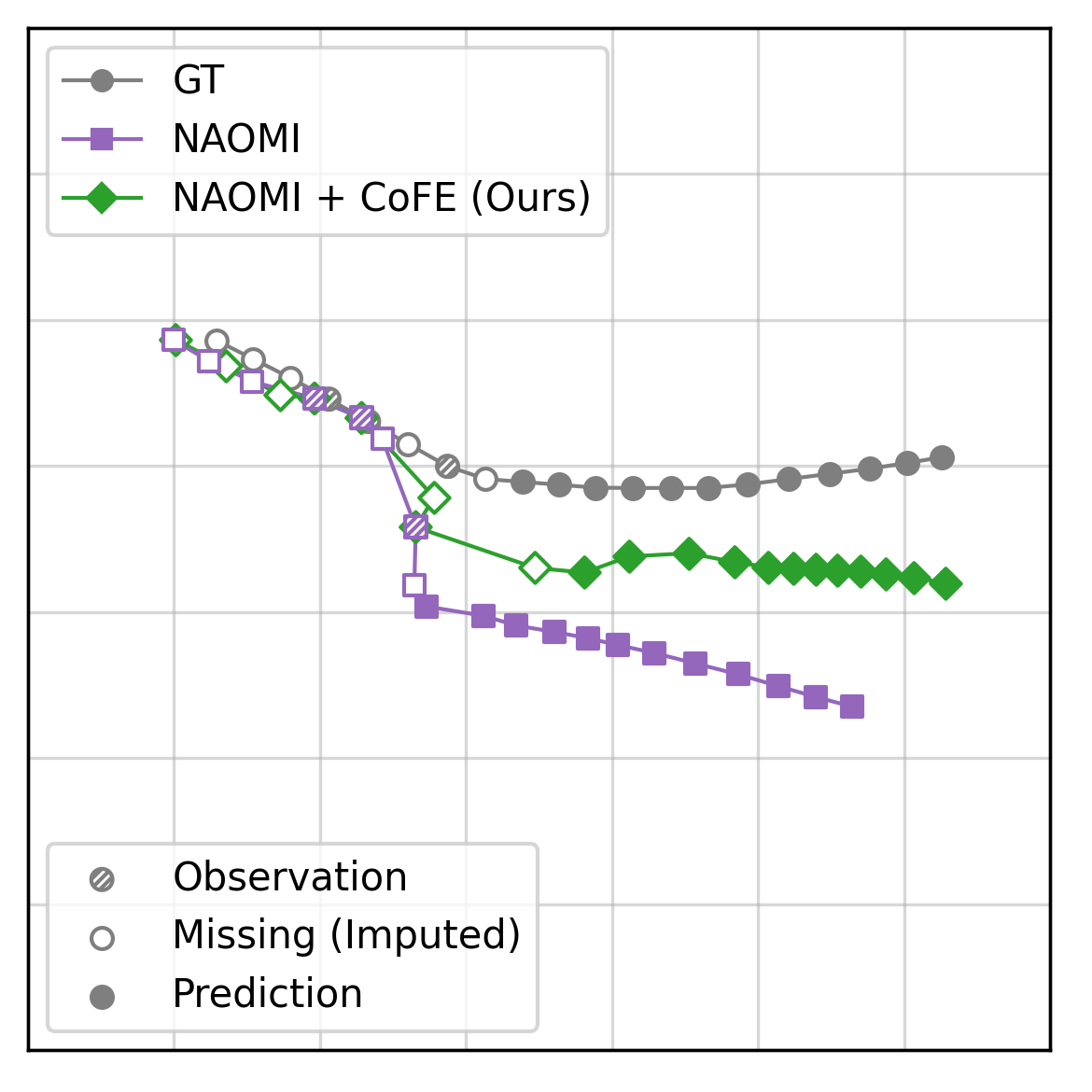}   
    }

    \caption{\textbf{Qualitative Results}: We demonstrate the effectiveness of CoFE, when applied to different imputation methods, using SGNet~\cite{sgnet} for prediction. Grid lines represent $1m$ in distance in ground plane.    
    }
    % \vspace{-0.3cm}%
    \label{fig:example_of_trajpred}
\end{figure*}

\begin{table*}[t]
\vspace{2mm}
\definecolor{lightskyblue}{rgb}{0.83, 0.91, 0.98}
\newcommand{\cboxblue}[1]{{\setlength{\fboxsep}{2pt}\colorbox{white}{\setlength{\fboxsep}{2pt}#1}}}
\newcommand{\cboxred}[1]{{\setlength{\fboxsep}{2pt}\colorbox{green!15}{\setlength{\fboxsep}{2pt}#1}}}

%\caption{ADE / FDE for each fold and approach tested on T2FPV-ETH dataset. The better result between using CoFE and not is \textbf{bolded}. The best result for each fold and prediction method is \cboxblue{blue}. The overall best performance for each prediction method is \cboxred{green}.}
\caption{ADE / FDE for each fold and approach tested on T2FPV-ETH dataset. The better result between using CoFE and not is \textbf{bolded} and the overall best performance for each prediction method is \cboxred{green}. Lower is better.}
\label{tab:ade_fold}

% \vspace{-2mm}
\centering

\newcommand{\w}{\cellcolor{white}}
\newcommand{\best}{\cellcolor{green!15}}

\begin{tabular}{c|cc|ccccc|c}
\multicolumn{9}{r}{Unit: meter} \\
\toprule[1pt]
 Traj. Prediction & Imputation & CoFE (ours) & ETH & Hotel & Univ & Zara1 & Zara2 & Avg \\

\midrule[1pt]
 \multirow{6}{*}{VRNN~\cite{vrnn}} & Linear-interp & - & \textbf{\w{1.35} / \w{2.00}} & 1.30 / 1.73 & 2.24 / 2.89 & 1.14 / 1.68 & 1.54 / 2.10 & 1.51 / 2.08 \\
 & Linear-interp & \checkmark & 1.52 / 2.35 & \textbf{\w{1.06} / \w{1.53}} & \textbf{1.65 / 2.10} & \textbf{1.06 / 1.63} & \textbf{1.27 / 1.62} & \textbf{1.31 / 1.84} \\
 % \cline[2-9]
 \cmidrule[0.5pt]{2-9}
 & Smooth~\cite{retrack} & - & 2.25 / 3.75 & 1.53 / 2.36 & 2.17 / 3.00 & 1.26 / 1.95 & 1.58 / 2.22 & 1.76 / 2.65 \\
 & Smooth~\cite{retrack} & \checkmark & \textbf{1.57 / 2.46} & \textbf{1.08 / 1.55} & \textbf{1.77 / 2.31} & \textbf{1.00 / 1.44} & \textbf{1.31 / 1.68} & \textbf{1.35 / 1.89} \\
 \cmidrule[0.5pt]{2-9}
 & NAOMI~\cite{naomi19} & - & \textbf{1.46 / 2.29} & 1.59 / 2.17 & 1.83 / 2.31 & 0.96 / 1.57 & 1.13 / 1.49 & 1.39 / 1.97 \\
 & NAOMI~\cite{naomi19} & \checkmark & 1.54 / 2.34 & \textbf{1.09 / 1.55} & \textbf{\w 1.58 / 1.97} & \textbf{\w{0.92} / \w{1.39}} & \textbf{\w{1.11} / \w{1.39}} & \best \textbf{1.25 / 1.73} \\
 \midrule[1pt]
 \multirow{6}{*}{A-VRNN~\cite{acvrnn}} & Linear-interp & - & \textbf{\w{1.39} / \w{2.04}} & 1.31 / 1.75 & 2.26 / 3.00 & 1.04 / 1.40 & 1.47 / 1.93 & 1.49 / 2.03 \\
 & Linear-interp & \checkmark & 1.47 / 2.18 &  \textbf{1.16 / 1.72} & \textbf{\w{1.54} / \w{1.88}} & \textbf{1.03 / 1.48} & \textbf{1.31 / 1.69} & \textbf{1.30 / 1.79} \\
 \cmidrule[0.5pt]{2-9}
 & Smooth~\cite{retrack} & - & 1.77 / 3.11 & 1.36 / 1.81 & 2.27 / 3.34 & 1.18 / 1.72 & 1.59 / 2.07 & 1.63 / 2.41 \\
 & Smooth~\cite{retrack} & \checkmark & \textbf{1.69 / 2.71} & \textbf{\w{1.10} / \w{1.59}} & \textbf{1.76 / 2.38} & \textbf{1.06 / 1.59} & \textbf{1.28 / 1.62} & \textbf{1.38 / 1.98} \\
 \cmidrule[0.5pt]{2-9}
 & NAOMI~\cite{naomi19} & - & \textbf{1.44 / 2.17} & 1.66 / 2.30 & 1.82 / 2.25 & \textbf{\w{0.83} / \w{1.24}} & \textbf{\w{1.09} / \w{1.39}} & 1.37 / 1.87 \\
 & NAOMI~\cite{naomi19} & \checkmark & 1.49 / 2.18 & \textbf{1.16 / 1.66} & \textbf{1.54 / 1.91} & 0.88 / 1.31 & 1.16 / 1.49 & \best \textbf{1.25 / 1.71} \\
\midrule[1pt]
 \multirow{6}{*}{SGNet~\cite{sgnet}} & Linear-interp & - & 1.43 / 1.97 & 0.72 / 1.00 & 1.48 / 1.73 & 0.58 / 0.79 & 0.78 / 0.91 & 1.00 / 1.28 \\
 & Linear-interp & \checkmark & \textbf{0.98 / 1.32} & \textbf{\w{0.59} / \w{0.76}} & \textbf{1.23 / 1.48} & \textbf{0.55 / 0.76} & \textbf{0.73 / 0.86} & \textbf{0.82 / 1.04} \\
 \cmidrule[0.5pt]{2-9}
 & Smooth~\cite{retrack} & - & 1.06 / 1.45 & 0.73 / 1.04 & 1.45 / 1.68 & 0.57 / 0.78 & 0.79 / 0.93 & 0.92 / 1.18 \\
 & Smooth~\cite{retrack} & \checkmark & \textbf{1.03 / 1.41} & \textbf{0.61 / 0.80} & \textbf{1.28 / 1.54} & \textbf{0.56 / 0.77} & \textbf{0.74 / 0.86} &  \textbf{0.84 / 1.08} \\
 \cmidrule[0.5pt]{2-9}
 & NAOMI~\cite{naomi19} & - & \textbf{\w{0.90} / \w{1.28}} & 0.78 / 0.97 & 1.21 / 1.43 & \textbf{\w{0.50} / \w{0.69}} & 0.72 / 0.84 & 0.82 / 1.04 \\
 & NAOMI~\cite{naomi19} & \checkmark & 0.99 / 1.40 & \textbf{0.59 / 0.78} & \textbf{\w{1.16} / \w{1.39}} & 0.51 / 0.70 & \textbf{\w{0.67} / \w{0.79}} & \best \textbf{0.78 / 1.01} \\
\bottomrule[1pt]
\end{tabular}

\vspace{-2mm}
\end{table*}
\begin{table}[t]
% \vspace{0.4cm}
\caption{Ablation study on CoFE applied to SGNet with linear interpolation.}
\label{tab:ablation}
\definecolor{lightgray}{gray}{0.95}
\centering
\begin{tabular}{c|c|cc|c}
\toprule
 Algorithm & CoFE & Train E2E & Impute Only & ADE / FDE \\
\midrule
\rowcolor{lightgray}
 \cellcolor{white} &  - & - & - & 1.00 / 1.28 \\
%  % \multirow{4}{*}{\rotatebox{90}{SGNet~\cite{sgnet}+CoFE}} & \checkmark & - & ??? / ??? \\
 %\midrule
  \cmidrule[0.5pt]{2-5}
 & \checkmark & No & No & 1.11 / 1.43 \\
 \rowcolor{lightgray}
 \cellcolor{white} & \checkmark & No & Yes & 0.98 / 1.27 \\
 & \checkmark & Yes & No & 0.94 / 1.22 \\
%\multirow{-5}{*}{\rotatebox{90}{SGNet~\cite{sgnet}}} & \rowcolor{lightgray} \checkmark & Yes & Yes & \textbf{0.82} / \textbf{1.04} \\
\rowcolor{lightgray}
 \cellcolor{white}\multirow{-5}{*}{{SGNet~\cite{sgnet}}} & \checkmark & Yes & Yes & \textbf{0.82} / \textbf{1.04} \\
\bottomrule
\end{tabular}
% \vspace{-3mm}
\end{table}

\section{EXPERIMENTS}
\label{sec:experiments}

\subsection{Experimental Setup}
\label{sec:setup}

We implemented several representative approaches on the ETH/UCY trajectory prediction task. We selected these algorithms as they stood out along several key techniques common in human trajectory prediction: variational prediction (VRNN~\cite{vrnn}), social awareness (A-VRNN~\cite{acvrnn}), and goal conditioning (SGNet~\cite{sgnet}).

For data imputation, we incorporated three commonly used approaches. We selected linear interpolation (``Linear-interp''), a simple but powerful approach used as part of many recent works, such as ~\cite{weng2022whose}. We also selected double exponential smoothing (``Smooth''), used in ~\cite{retrack}, a more complex baseline that better handles dynamic trends in the sequence. Finally, we incorporated NAOMI~\cite{naomi19}, which is a recent SOTA approach leveraging deep learning.

% This approach led us to select VRNN~\cite{vrnn}, A-VRNN~\cite{acvrnn}, and SGNet~\cite{sgnet} as our initial algorithms to examine.

%\soonmin{(Do we really want to mention it?)}Note that A-VRNN is an ablation of AC-VRNN, which adds in goal conditioning in a somewhat similar manner to SGNet, but ultimately performs worse in reported results; hence, we did not include it in our study. 

% \soonmin{(As we talked about this in Sec 2, we don't have to say this twice.)}While there are performers on the ETH/UCY benchmark leaderboard which report performing better than SGNet, as discussed in Section \ref{sec:rel_work}, they generally rely on additional input modalities beyond the scope of our task. Additional approaches would certainly be interesting to incorporate and study as well, but we leave this as future work extending beyond the scope of this study. \soonmin{Seems wordy?}

\subsection{Evaluation Procedure}
\label{metrics}

As in Social GAN~\cite{sgan}, we evaluate trajectory predictions using a leave-one-out approach. For each of the five folds, models are trained and validated on data from four of them at a time. Then, the best model according to validation performance is tested on the entirety of the held-out fold.

We train NAOMI~\cite{naomi19} separately, following the author's procedure, once per fold. Then, for each combination of imputation techniques and prediction algorithms, we train one prediction model utilizing CoFE and one model without (i.e., just using the $\mathbf{\tilde{X}_{d_i}^{hist}}$ outputs from the imputation). 

In the field of trajectory prediction, and especially for ETH/UCY, the most commonly used metrics are Average Displacement Error (ADE) and Final Displacement Error (FDE). These metrics can be easily computed on a per-agent basis, for ground truth future track $\mathbf{X_i^{fut}}$  and predicted future track $\mathbf{\hat{X}_i^{fut}}$, for each agent $i$ in the scene. The $L2$-distance at each time $t$ from $t=T_{obs+1}$ to $t=T_{pred}$ is taken between $\mathbf{X_i^{t}}$ and $\mathbf{\hat{X}_i^{t}}$; ADE is the average of these distances, while FDE is the final distance.

% However, since we consider detection and tracking to be a core part of the challenge posed by this dataset, our metrics must extend beyond ADE/FDE and incorporate a sense of precision and recall regarding which agents should have been observed and predicted in the first place. This is made even more challenging by the fact that there is no guarantee of alignment of agent IDs between the observed tracks and ground truth tracks, even if the precision and recall is perfect. To account for these issues, we use a variant of mean Average Precision~\cite{map} (mAP) for trajesctory prediction, using ADE as a stand-in for both confidence and match quality.

%Note that all approaches we consider utilize a ``best-of-$K$'' sample prediction strategy. \soonmin{(We have to mention that this is a standard evaluation protocol and we just follow it. Because it might sound hacky.)} This accounts for the fact that there are multiple socially valid predictions for agents in a scene, so feasible (but ultimately incorrect) predictions should be punished less. 
% As a result, ADE and FDE are computed in a $K$-to-one manner, for each sample, and are reduced to their minimum value before being passed to downstream mAP computation. 
We note that there are other metrics which could be utilized, including collision rate, social comfort level, path complexity, and many more \cite{coresrn}. We chose to focus on the core metrics of the tasks at hand, but suggest that future work in applying these metrics could provide helpful new insight. 
% \section{RESULTS}
\subsection{Results}
% \label{sec:results}

We conducted extensive experimentation to assess CoFE's performance when combined with the various imputation and prediction approaches. As shown in Table \ref{tab:ade_fold}, adding the CoFE module is quite effective. When considering the average performance over all dataset folds, all combinations of imputation and prediction algorithms which use CoFE are better than the corresponding versions which bypass it. 
%This suggests that CoFE is effective at capturing underlying patterns in the data that are not captured by the individual imputation and prediction algorithms alone.
Furthermore, without CoFE, the average performance is highly variable, dependent on the choice of imputation approach and prediction method. However, with CoFE, these differences become much less pronounced. 
Note that although performance on most folds in most cases is quite good, CoFE appears to not be as effective on ETH. We suspect that this is because, as shown in Table \ref{tab:simple_stats}, there are only a small number of detected tracks in the first place (60), so performance is more variable and sensitive to any individual prediction's error.

To gain further insight into CoFE's performance, we performed qualitative analyses. As seen in Figure \ref{fig:example_of_trajpred}-(b), the imputation approach (NAOMI~\cite{naomi19}) trusts surrounding points in the data, performing an extrapolation and thus, does not effectively capture the FPV errors. When paired with CoFE, the approach is more effective at capturing underlying temporal and spatial patterns in the data, correcting the FPV errors which results in better downstream prediction. 

We also performed an ablation study, shown in Table \ref{tab:ablation}, to assess aspects of our design choices. We focused on SGNet~\cite{sgnet} combined with linear interpolation for the study, and found clearly that the E2E training was vital in obtaining top performance. We further find that focusing on imputation only in the prediction phase (i.e., replacing non-imputed points in $\mathbf{\hat{X}_{d_i}^{hist}}$ as described in Section \ref{sec:training}) also has a significant effect in improving performance.
\section{FUTURE WORK}
\label{limitations}
Although SEANavBench is a high-fidelity environment, we do note that further effort in improving its realism could be useful. Realism could be enhanced not just by increasing the 3D-modeling asset and animation qualities, but also by further improving alignment between the reproduced scenery and the original locations. 

Additionally, for associating D\&T tracks with their corresponding GT tracks, we relied on Hungarian matching on our tracking output directly. This decreased the number of correctly matched trajectories, due to identity association errors of detections. Incorporating affinity-based techniques from \cite{weng2022whose} or performing the full re-tracking algorithm from \cite{retrack} could be a promising way to even further reduce FPV errors.

\section{CONCLUSION}
In existing work, pedestrian trajectory prediction has been mainly studied under a complete information assumption. In this paper, we introduce a first-person view trajectory prediction problem where agents need to make predictions based on partial, imprecise information. To promote this research direction, we present T2FPV, a method for generating high-fidelity egocentric datasets for pedestrian navigation by leveraging existing real-world trajectory datasets. 
In this setting, FPV-specific errors arise due to imperfect detection and tracking, occlusions, and FOV limitations of the camera. To address these errors, we propose CoFE, a module that further refines imputation of missing data in an end-to-end manner with trajectory forecasting algorithms. Our method reduces the impact of such FPV errors on downstream prediction performance, decreasing displacement error by 11.73\%, averaging over all combinations of imputation techniques and prediction approaches tested. We also show that E2E training of CoFE is essential in achieving this performance increase. 
Our constructed T2FPV-ETH dataset provides a benchmark for human trajectory prediction from detection and tracking results, which is a more natural and realistic setting. Therefore, we argue that incorporating such realism throughout the perception pipeline is an important direction to move toward in enabling robots to navigate in the real world.

%\addtolength{\textheight}{-12cm}

%\section*{ACKNOWLEDGMENT}

%%%%%%%%%%%%%%%%%%%%%%%%%%%%%%%%%%%%%%%%%%%%%%%%%%%%%%%%%%%%%%%%%%%%%%%%%%%%%%%%

\bibliographystyle{IEEEtran} % We choose the "IEEEtran.bst" reference style
\bibliography{IEEEabrv} % Entries are in the "IEEEabrv.bib" file

\end{document}